\documentclass{article}

\usepackage[preprint]{neurips_2025}

\usepackage[utf8]{inputenc} 
\usepackage[T1]{fontenc}    
\usepackage{hyperref}       
\usepackage{url}            
\usepackage{booktabs}       
\usepackage{amsfonts}       
\usepackage{nicefrac}       
\usepackage{microtype}      
\usepackage{xcolor}         
\usepackage{graphicx} 
\usepackage{amsmath} 
\usepackage{multirow}
\usepackage{wrapfig}
\usepackage{placeins}
\usepackage{subcaption}
\usepackage{tikz}
\usepackage{colortbl}
\usepackage{enumitem}
\definecolor{customteal}{HTML}{1F4040}

\usetikzlibrary{
    shapes.geometric,
    arrows.meta,
    positioning,
    backgrounds,
    shadows.blur
}

\title{Amortized Latent Steering: Low-Cost Alternative to Test-Time Optimization}

\author{%
 Nathan Egbuna \quad Saatvik Gaur \quad Kevin Zhu \\
 \textbf{Sunishchal Dev} \quad \textbf{Ashwinee Panda} \quad \textbf{Maheep Chaudhary}\thanks{Project Lead} \\[0.3em]
 Algoverse AI Research \\
 \texttt{negbuna26@andover.edu, maheepchaudhary.research@gmail.com} \\[0.3em]
}

\begin{document}

\maketitle

\begin{abstract}
Test-time optimization remains impractical at scale due to prohibitive inference costs---techniques like iterative refinement and multi-step verification can require $10$--$100\times$ more compute per query than standard decoding. Latent space test-time optimization methods like LatentSeek offer a more direct approach by steering hidden representations, but still demand expensive per-query optimization loops with multiple backward passes. We propose \textsc{Amortized Latent Steering (ALS)}, which collapses this iterative optimization into a single offline-computed vector applied at constant cost during inference. \textsc{ALS} computes the mean difference between hidden states from successful versus unsuccessful generations, then uses this direction to calibrate the model's hidden representations: when decoding drifts away from the success manifold, \textsc{ALS} nudges activations back toward it. Across GSM8K and MATH-500 benchmarks, \textsc{ALS} achieves $2$--$5\times$ speed-up over iterative methods while matching or surpassing greedy Chain-of-Thought and Self-Consistency baselines, yielding up to 101\% improvement in efficiency--accuracy trade-off. These results show that much of latent optimization's benefit can be captured offline, making sophisticated reasoning techniques viable for production deployment. Code is available at~\href{https://github.com/negbuna/ALS}{\texttt{https://github.com/negbuna/ALS}}.

\begin{figure}[h]
    \centering
    \begin{subfigure}[b]{0.49\textwidth}
        \centering
        \includegraphics[width=0.95\linewidth]{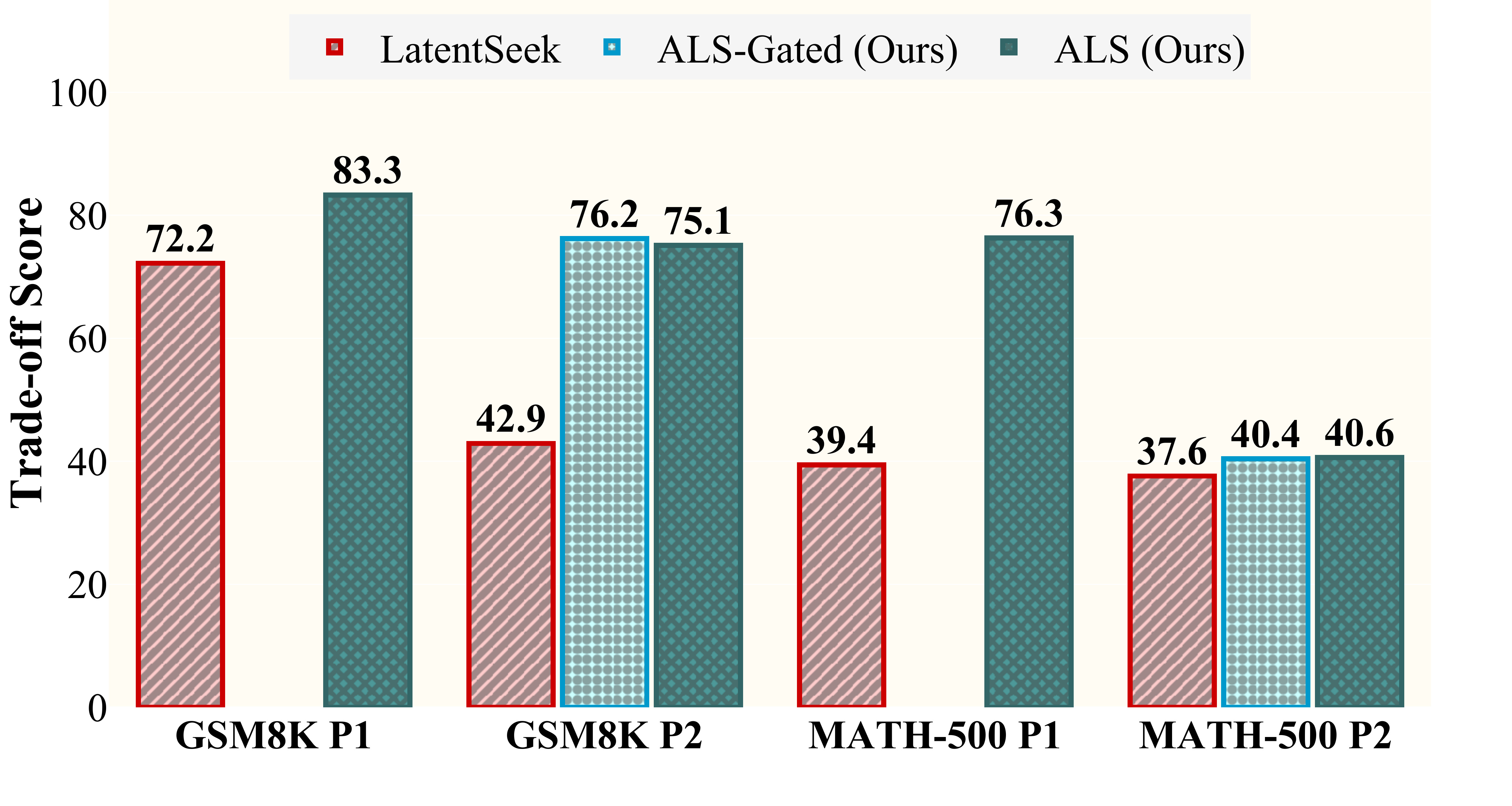}
        \caption{Qwen-2.5-7B-Instruct}
        \label{fig:qwen}
    \end{subfigure}
    \hfill
    \begin{subfigure}[b]{0.49\textwidth}
        \centering
        \includegraphics[width=0.95\linewidth]{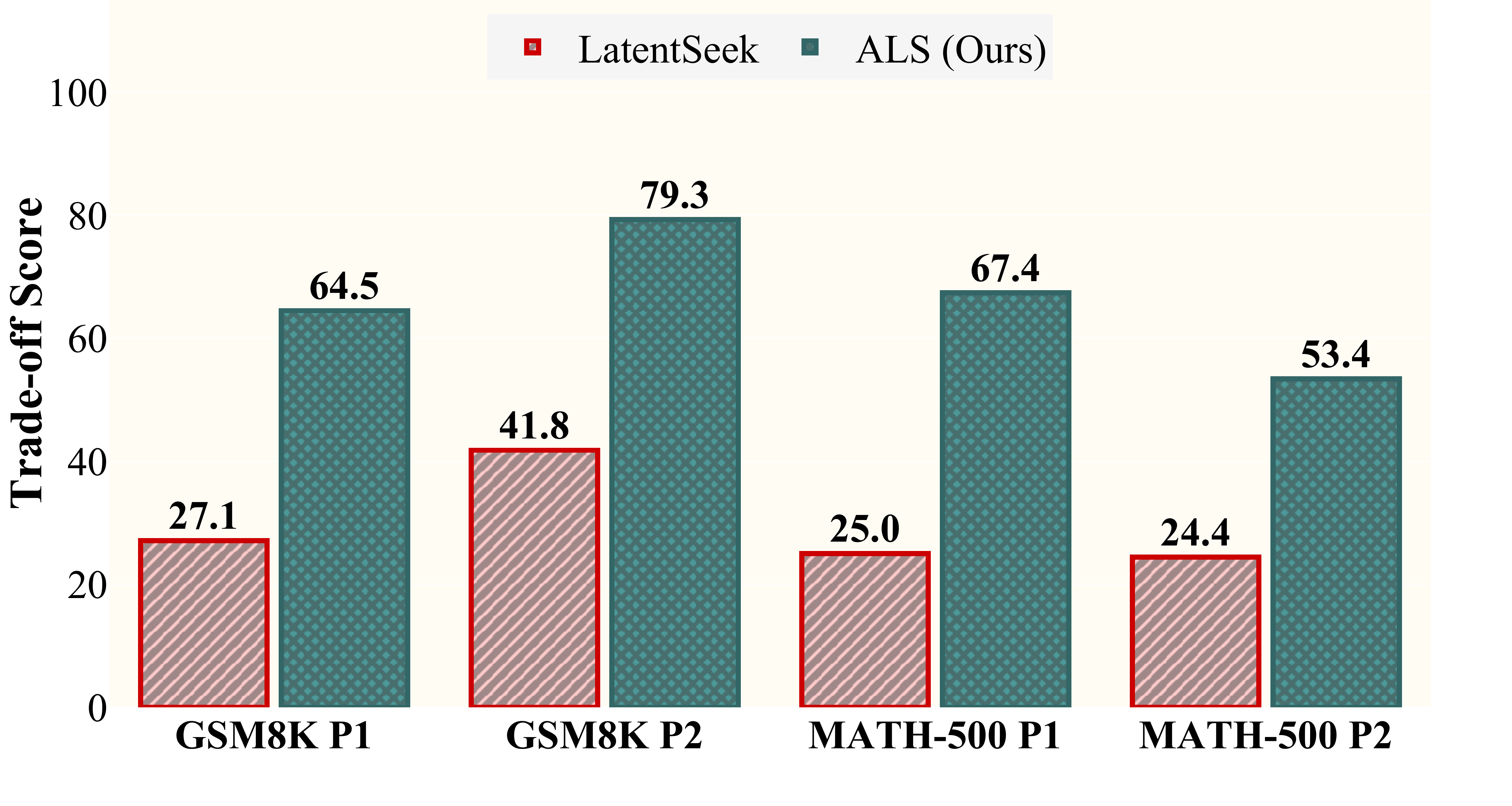}
        \caption{Llama-3.1-8B-Instruct}
        \label{fig:llama}
    \end{subfigure}
    \caption{\textbf{\textsc{ALS} consistently outperforms LatentSeek across all evaluation settings while matching or exceeding Chain-of-Thought performance.} Results show efficiency--accuracy trade-offs on GSM8K and MATH-500 for both Qwen-2.5-7B and Llama-3.1-8B models.}
    \label{fig:model_results}
\end{figure}

\end{abstract}

\section{Introduction}

Test-time optimization (TTO) methods require $10$--$100\times$ more compute than standard decoding, making them impractical for production deployment. Even sophisticated latent-space methods like LatentSeek demand expensive per-query optimization loops with multiple backward passes.

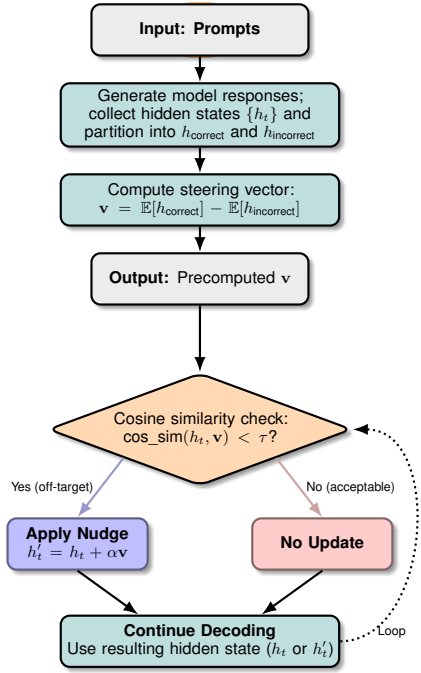
\begin{wrapfigure}[29]{r}{0.6\textwidth} 
  \centering
  \resizebox{0.712\linewidth}{!}{
        \begin{tikzpicture}[
    scale=0.8,
    node distance=0.5 cm,
    font=\sffamily\small
]

\colorlet{bgColor}{gray!5}
\colorlet{borderColor}{orange!70}
\colorlet{processColor}{teal!25}
\colorlet{decisionColor}{orange!30}
\colorlet{yesColor}{blue!25}
\colorlet{noColor}{red!20}

\tikzset{
    base/.style={draw, very thick, rounded corners=5pt, blur shadow={shadow blur steps=5}, text centered},
    process/.style={base, fill=processColor, text width=5.2cm, minimum height=1cm},
    decision/.style={base, diamond, aspect=2.7, fill=decisionColor, text width=4cm, inner sep=0.5pt},
    io/.style={base, fill=gray!15, text width=4cm, minimum height=1cm},
    outcome/.style={base, text width=2.5 cm, minimum height=1cm},
    arrow/.style={->, >=Latex, very thick}
}

\node (prompts) [io] {\textbf{Input: Prompts}};
\node (generate) [process, below=of prompts] {Generate model responses; collect hidden states $\{h_t\}$ and partition into $h_{\text{correct}}$ and $h_{\text{incorrect}}$};
\node (compute) [process, below=of generate] {Compute steering vector: $\mathbf{v} = \mathbb{E}[h_{\text{correct}}] - \mathbb{E}[h_{\text{incorrect}}]$};
\node (output) [io, below=of compute] {\textbf{Output:} Precomputed $\mathbf{v}$};

\node (check) [decision, below=1.3cm of output] 
{Cosine similarity check:\\ $\text{cos\_sim}(h_t, \mathbf{v}) < \tau$?};
\node (nudge) [outcome, fill=yesColor, draw=yesColor!50!black, 
    below left=0.6 cm and 1 cm of check.south] 
{\textbf{Apply Nudge} \\ $h_t' = h_t + \alpha \mathbf{v}$};

\node (no_nudge) [outcome, fill=noColor, draw=noColor!50!black, 
    below right=0.6 cm and 1 cm of check.south] 
{\textbf{No Update}};
\node (decode) [process, below=2.5cm of check] 
{\textbf{Continue Decoding} \\ Use resulting hidden state ($h_t$ or $h_t'$)};

\draw[arrow] (prompts) -- (generate);
\draw[arrow] (generate) -- (compute);
\draw[arrow] (compute) -- (output);
\draw[arrow] (output) -- (check);
\draw[arrow, color=yesColor!80!black] (check.south west) -- (nudge.north) 
    node[midway, left, font=\sffamily\scriptsize, text=black]{Yes (off-target)};
\draw[arrow, color=noColor!80!black] (check.south east) -- (no_nudge.north) 
    node[midway, right, font=\sffamily\scriptsize, text=black]{No (acceptable)};
\draw[arrow] (nudge.south) -- ([xshift=-1.5cm]decode.north);
\draw[arrow] (no_nudge.south) -- ([xshift=1.5cm]decode.north);

\draw[arrow, dotted] 
  (decode.east) to[out=0, in=0, looseness=1.1] 
  node[pos=0.125, right, font=\sffamily\scriptsize]{Loop} 
  (check.east);

\begin{scope}[on background layer]
    \node (container) [
        draw=borderColor, 
        fill=bgColor, 
        rounded corners=10pt,
        thick,
        inner sep=15pt
    ] {};
\end{scope}

\end{tikzpicture}
  }
  \caption{\textsc{ALS} workflow. Offline, the model computes a steering vector to nudge hidden states $\{h_t\}$ toward successful reasoning trajectories.}
  \label{fig:als_workflow}
\end{wrapfigure}

We propose \textsc{Amortized Latent Steering (ALS)}, inspired by causal intervention techniques that identify which latent representations causally influence model outputs. Rather than iteratively optimizing during inference, ALS pre-computes steering directions by analyzing the causal relationship \cite{liu2023towards, geiger2025causal} between hidden states and reasoning success. The key insight is that we can intervene on the model's internal representations using directions that causally promote correct reasoning, applying these interventions at constant cost during inference.

\textsc{ALS} achieves dramatic efficiency gains, with $2$--$5\times$ speed-up over iterative methods while matching or surpassing greedy Chain-of-Thought (CoT) and Self-Consistency baselines across GSM8K and MATH-500. Most significantly, on the challenging MATH-500 benchmark, \textsc{ALS} yields up to 101\% improvement in efficiency--accuracy trade-off, demonstrating that the computational benefits of latent optimization can be fully amortized offline without performance degradation.

Our contributions are threefold: 
\begin{enumerate}[leftmargin=*,topsep=2pt,itemsep=2pt]
    \item We propose a computationally practical latent steering method that amortizes expensive TTO into a single offline-computed vector.
    \item ALS provides superior efficiency--accuracy trade-offs and up to 101\% improvement on challenging benchmarks.
    \item We provide insights on relationship between steering strength and model architecture for effective intervention.
\end{enumerate}

\section{Related Work}
\label{rel}

TTO for reasoning spans a spectrum from expensive iterative methods to lightweight interventions. Optimization-heavy approaches like LatentSeek \cite{latentseek} and Fractional Reasoning \cite{fractionalreasoning} iteratively adjust hidden states at test time but require expensive query-specific operations and substantial computational overhead. Notably, LatentSeek itself was only evaluated on mathematical reasoning benchmarks, such as GSM8K, MATH-500, and AIME-2024, making our focus on GSM8K and MATH-500 directly comparable and consistent with prior work. 

Linear intervention methods including COCONUT \cite{coco} and activation steering \cite{activationsteering, repeng} show that hidden state modifications can shift model behavior, but typically require extensive task-specific tuning and are better suited for synthetic QA or stylistic control rather than complex symbolic reasoning tasks.

\textsc{ALS} bridges these approaches by using offline computation to derive a single steering vector from successful vs. unsuccessful trajectories, then applying lightweight linear interventions during inference. Unlike optimization-heavy methods, \textsc{ALS} requires no backward passes or large online overhead; unlike existing activation steering, it specifically targets symbolic reasoning accuracy through success/failure trajectory differences. This design enables \textsc{ALS} to achieve the accuracy benefits of iterative optimization and surpass standard decoding baselines while maintaining constant-time inference overhead.

\section{Method}

Test-time optimization methods like LatentSeek achieve strong reasoning performance but require expensive iterative backward passes for each query. We propose \textsc{Amortized Latent Steering (ALS)}, which collapses this per-query optimization into a single offline computation, enabling constant-cost inference while preserving the benefits of latent-space intervention.

\subsection{Steering Vector Construction}

\textsc{ALS} collapses iterative TTO into a single pre-computed steering vector. To compute the vector, we generate model outputs from 1,000 GSM8K \cite{gsm8k} and 500 MATH \cite{math500} examples (disjoint from evaluation) and extract $h_t$, the hidden state at token $t$, at the penultimate layer for each generation, consistent with prior latent intervention work \cite{coco}. 

We extract hidden states from the penultimate layer as this layer contains high-level semantic representations while remaining sufficiently close to the output to influence generation. We also explored injecting steering vectors at earlier and deeper layers. However, modifying hidden states outside the penultimate layer consistently caused attention mask mismatches in the \texttt{transformers} library due to residual connections expecting tensors with fixed shapes. This made evaluation unstable, so we restricted our analysis to the penultimate layer, which was both stable and semantically rich.

The combination of outputs yields two distributions of $h_t$ (correct vs. incorrect). Correctness is determined by automatic answer verification against ground-truth, following LatentSeek. For JSON prompts, exact format validity is also required. The steering vector is defined as
\[
\mathbf{v} = \mathbb{E}[h_{\text{correct}}] - \mathbb{E}[h_{\text{incorrect}}]
\]
capturing the latent difference between successful and failed trajectories.

\subsection{Test-Time Intervention}

At test time, we monitor cosine similarity between the current hidden state $h_t$ and the steering vector $\mathbf{v}$ at each token. When similarity falls below threshold $\tau = 0.1$, we apply a lightweight additive update:
\[
h_t' = h_t + \alpha \mathbf{v},
\]
where $\alpha$ controls steering strength. The threshold $\tau = 0.1$ was selected through validation on 200 held-out examples, balancing intervention frequency (applied to $\sim$30\% of tokens) with performance. Higher values led to over-steering and degraded accuracy, while lower values missed correction opportunities. 

This monitoring and intervention process requires only vector operations with no backward passes, adding negligible computational overhead comparable to standard decoding. Unlike LatentSeek, which requires $O(k \cdot B)$ operations at test time for $k$ backward passes of cost $B$, \textsc{ALS} reduces this to $O(1)$ per token---only a cosine similarity computation and potential vector addition per step.

\subsection{Structured Output Variant}

We explored a gated variant (\textsc{ALS-Gated}) for structured outputs. At each decoding step, \textsc{ALS-Gated} parses the partial generation to check whether it remains valid JSON. If adding the next token would break the structure (i.e., cause a parsing error), the steering update is suppressed by setting $\alpha=0$ for that step. This gating mechanism ensures that latent interventions do not corrupt required output formats while still allowing steering to improve reasoning quality within the valid schema.

\section{Experimention and Results}

We evaluate \textsc{ALS} on mathematical reasoning benchmarks using two instruction-tuned language models, comparing against strong baselines representing different points in the efficiency--accuracy spectrum.

\subsection{Experimental Setup}

\paragraph{Models and Datasets}

We evaluated \textsc{ALS} on two open source instruction-tuned models, Qwen-2.5-7B-Instruct and Llama-3.1-8B-Instruct, using the full GSM8K test set (1,319 examples) and MATH-500 dataset (500 examples). 
For each dataset, we consider two prompt formats, following the experimental setup of LatentSeek. In the first (P1), a free-form CoT prompt instructs the model to reason step by step in natural language before producing a final answer. In the second (P2), a structured JSON prompt requires the model to return its answer in a rigid schema \texttt{\{"thought process": "...", "final answer": "..."\}}, introducing syntactic constraints in addition to correctness requirements.

\paragraph{Baselines and Evaluation}

Baselines include greedy CoT decoding \cite{cot}, Self-Consistency \cite{sc} with $k=5$ samples, and LatentSeek, representing standard decoding and optimization-heavy latent methods. All methods use identical random seeds and problem splits for reproducibility. 
All experiments used $1\times$ NVIDIA H200. LatentSeek on Llama-3.1-8B proved prohibitively expensive ($>$100s per MATH-500 example), so we report results on a stratified 224-example subset for those settings.
We evaluate using a trade-off score that balances accuracy and efficiency:

\begin{equation}
\label{eq:tradeoff}
\text{Trade-off} = \frac{\text{Accuracy} + (100 - \text{Normalized Time})}{2}
\end{equation}
where normalized time scales the slowest method to 100. Accuracy and average inference latency serve as evaluation metrics, with trade-off scores combining both for comparison under efficiency--accuracy constraints.

\begin{table*}[h!]
\small
\footnotesize
\centering
\caption{Trade-off, accuracy (Acc.), and average inference time (s) for \textsc{ALS} (best $\alpha$) and standard baselines. The best method is highlighted in bold, and the second best is underlined. Teal numbers indicate the percent change of \textsc{ALS} trade-off relative to CoT.}
\label{tab:main_results}
{\footnotesize
\begin{tabular}{@{}c c @{}c c c c | c c c@{}}
\toprule
& & & \multicolumn{3}{c}{\textbf{Qwen-2.5-7B-Instruct}} & \multicolumn{3}{c}{\textbf{Llama-3.1-8B-Instruct}} \\
\cmidrule(lr){4-6} \cmidrule(lr){7-9}
Dataset & Prompt & Method & Acc. & Time & Trade-off & Acc. & Time & Trade-off \\
\midrule
\multirow{6}{*}{MATH-500}
 & P1 & LatentSeek & 75.4 & 47.0 & 39.4 & 50.0 & 106.8 & 25.0 \\
 & & CoT & 76.0 & 9.9 & \underline{77.8} & 52.0 & 10.8 & \textbf{70.8} \\
 & & SC & 76.0 & 48.6 & 38.0 & 52.0 & 54.0 & 26.0 \\
 & \cellcolor{gray!8} & \cellcolor{gray!8}\textsc{ALS} (Ours) & \cellcolor{gray!8}91.0 & \cellcolor{gray!8}5.2 & \cellcolor{gray!8}\textbf{93.1} {\scriptsize \textcolor{teal}{(+19.7\%)}} & \cellcolor{gray!8}44.8 & \cellcolor{gray!8}10.6 & \cellcolor{gray!8}\underline{67.4} {\scriptsize \textcolor{teal}{(-4.8\%)}} \\
\cmidrule(lr){2-9}
 & P2 & LatentSeek$^\dagger$ & 56.0 & 43.1 & 37.6 & 48.7 & 116.5 & 24.4 \\
 & & CoT & 3.5 & 10.7 & \underline{41.7} & 5.0 & 11.5 & \underline{44.8} \\
 & & SC & 3.0 & 53.3 & 1.5 & 5.0 & 56.2 & 1.4 \\
 & \cellcolor{gray!8} & \cellcolor{gray!8}\textsc{ALS}-Gated (Ours) & \cellcolor{gray!8}3.0 & \cellcolor{gray!8}11.8 & \cellcolor{gray!8}40.4 & \cellcolor{gray!8}- & \cellcolor{gray!8}- & \cellcolor{gray!8}- \\
 & \cellcolor{gray!8} & \cellcolor{gray!8}\textsc{ALS} (Ours) & \cellcolor{gray!8}68.5 & \cellcolor{gray!8}2.2 & \cellcolor{gray!8}\textbf{83.8} {\scriptsize \textcolor{teal}{(+101.0\%)}} & \cellcolor{gray!8}16.0 & \cellcolor{gray!8}10.7 & \cellcolor{gray!8}\textbf{53.4} {\scriptsize \textcolor{teal}{(+19.2\%)}} \\
\midrule
\multirow{6}{*}{GSM8K} 
 & P1 & LatentSeek & 90.4 & 9.8 & 72.2 & 54.1 & 18.8 & 27.1 \\
 & & CoT & 91.0 & 4.3 & \textbf{85.5} & 54.0 & 4.5 & \textbf{74.8} \\
 & & SC & 91.0 & 21.3 & 45.5 & 54.0 & 19.8 & 36.0 \\
 & \cellcolor{gray!8} & \cellcolor{gray!8}\textsc{ALS} (Ours) & \cellcolor{gray!8}90.6 & \cellcolor{gray!8}5.1 & \cellcolor{gray!8}\underline{83.3} {\scriptsize \textcolor{teal}{ (-2.6\%)}} & \cellcolor{gray!8}51.4 & \cellcolor{gray!8}4.2 & \cellcolor{gray!8}\underline{64.5} {\scriptsize \textcolor{teal}{ (-13.8\%)}}\\
\cmidrule(lr){2-9}
 & P2 & LatentSeek & 85.8 & 13.3 & 42.9 & 83.6 & 30.6 & 41.8 \\
 & & CoT & 66.0 & 1.7 & \textbf{76.5} & 60.0 & 3.2 & \underline{78.4}  \\
 & & SC & 66.0 & 8.6 & 50.5 & 60.0 & 15.5 & 37.3 \\
 & \cellcolor{gray!8} & \cellcolor{gray!8}\textsc{ALS}-Gated (Ours) & \cellcolor{gray!8}72.0 & \cellcolor{gray!8}2.6 & \cellcolor{gray!8}\underline{76.2} {\scriptsize \textcolor{teal}{ (-0.4\%)}} & \cellcolor{gray!8}- & \cellcolor{gray!8}- & \cellcolor{gray!8}- \\
 & \cellcolor{gray!8} & \cellcolor{gray!8}\textsc{ALS} (Ours) & \cellcolor{gray!8}70.4 & \cellcolor{gray!8}2.7 & \cellcolor{gray!8}75.1 & \cellcolor{gray!8}68.8 & \cellcolor{gray!8}3.1 & \cellcolor{gray!8}\textbf{79.3} {\scriptsize\textcolor{teal}{(+1.1\%)}} \\
\bottomrule
\end{tabular}
}

\footnotesize{$^\dagger$ Evaluated on a 224-example stratified subset due to compute cost.}
\end{table*}


Table~\ref{tab:main_results} compares \textsc{ALS} against baselines across models, datasets, and prompt styles. Table results show best $\alpha$ per setting, as performance varies with hyperparameter $\alpha$, as expected (see Table~\ref{tab:ablations}).
On GSM8K, \textsc{ALS} variants consistently outperform LatentSeek and compete closely with CoT. On the more challenging MATH-500 dataset, \textsc{ALS} shows greater sensitivity to $\alpha$ but consistently outperforms both LatentSeek and CoT while significantly reducing inference time.
Notably, when \textsc{ALS} underperforms relative to CoT, the margins are modest (0.4--14\% decrease), but when \textsc{ALS} succeeds, the improvements are substantial (19--101\% increase), indicating that the method either maintains competitive performance or delivers significant gains.

\subsection{Task Difficulty and Model Sensitivity}

On GSM8K, \textsc{ALS} demonstrates strong performance parity with CoT across both models and prompt types. However, on the more challenging MATH-500 benchmark, \textsc{ALS} shows dramatic improvements, particularly for Qwen-2.5-7B. This suggests \textsc{ALS} provides greater benefits on harder reasoning tasks where latent steering can more effectively guide the model away from incorrect solution paths.

Cross-architectural analysis reveals distinct sensitivities to latent interventions. Qwen-2.5-7B consistently shows larger improvements from \textsc{ALS} steering, particularly on MATH-500 where accuracy jumps from 76\% (CoT) to 91\% (\textsc{ALS}) on free-form prompts. In contrast, Llama-3.1-8B shows more modest accuracy improvements but maintains the efficiency gains, suggesting architecture-specific differences in latent geometry and steering effectiveness.

The structured JSON format (P2) presents unique challenges, with both models showing degraded CoT performance compared to free-form prompts. However, \textsc{ALS} demonstrates remarkable robustness: on MATH-500 P2, \textsc{ALS} achieves 68.5\% accuracy compared to CoT's 3.5\%, a 65 percentage point improvement that translates to a 101\% trade-off score increase, showing \textsc{ALS} can maintain reasoning coherence even under syntactic constraints where standard decoding fails.

\subsection{Ablation Studies}

\begin{figure}[h!]
    \centering
    \includegraphics[width=0.7\linewidth]{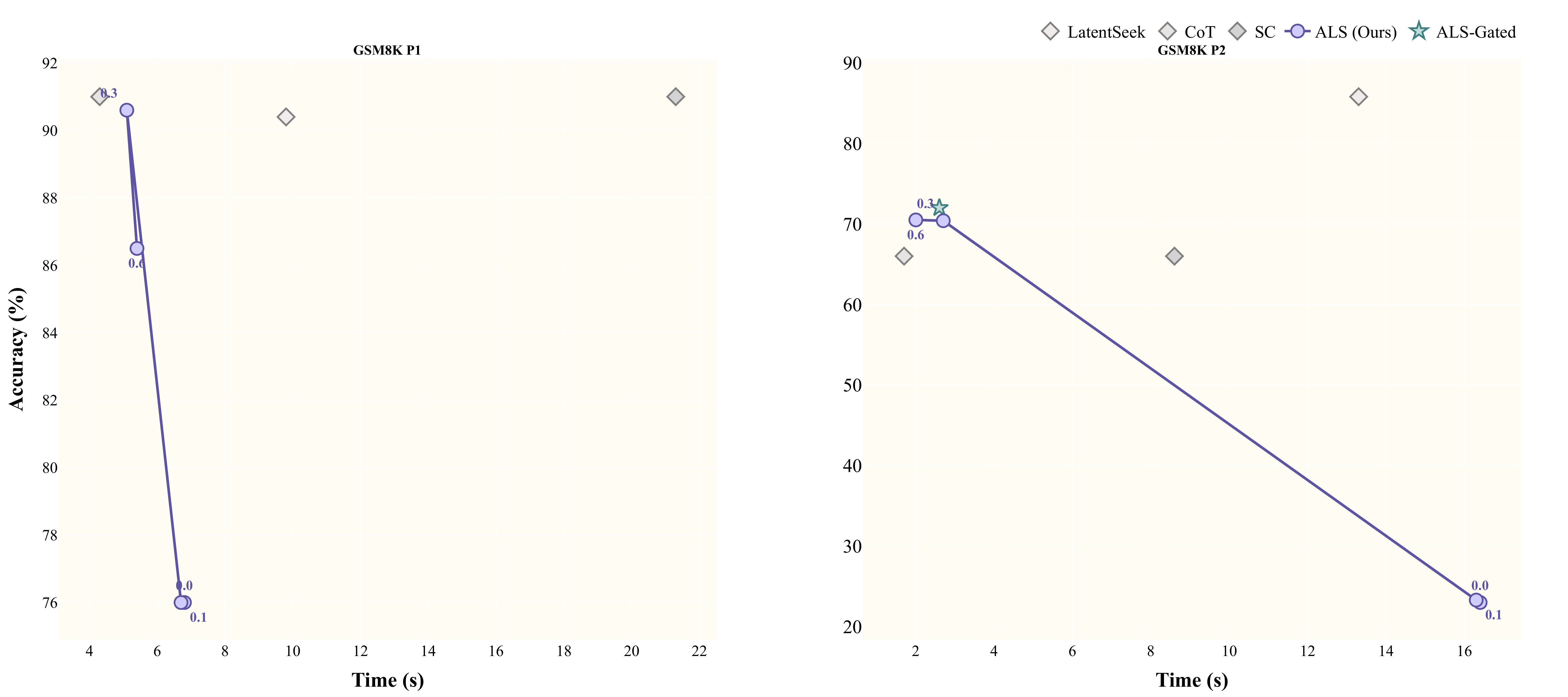}
    \caption{Pareto frontier for \textsc{ALS} on GSM8K. Each point corresponds to a steering strength $\alpha$, showing the trade-off between accuracy and inference time. The curve highlights how intermediate $\alpha$ values yield the best balance of efficiency and accuracy.}
    \label{fig:gsm_pareto}
\end{figure}

\begin{figure}[h!]
    \centering
    \includegraphics[width=0.7\linewidth]{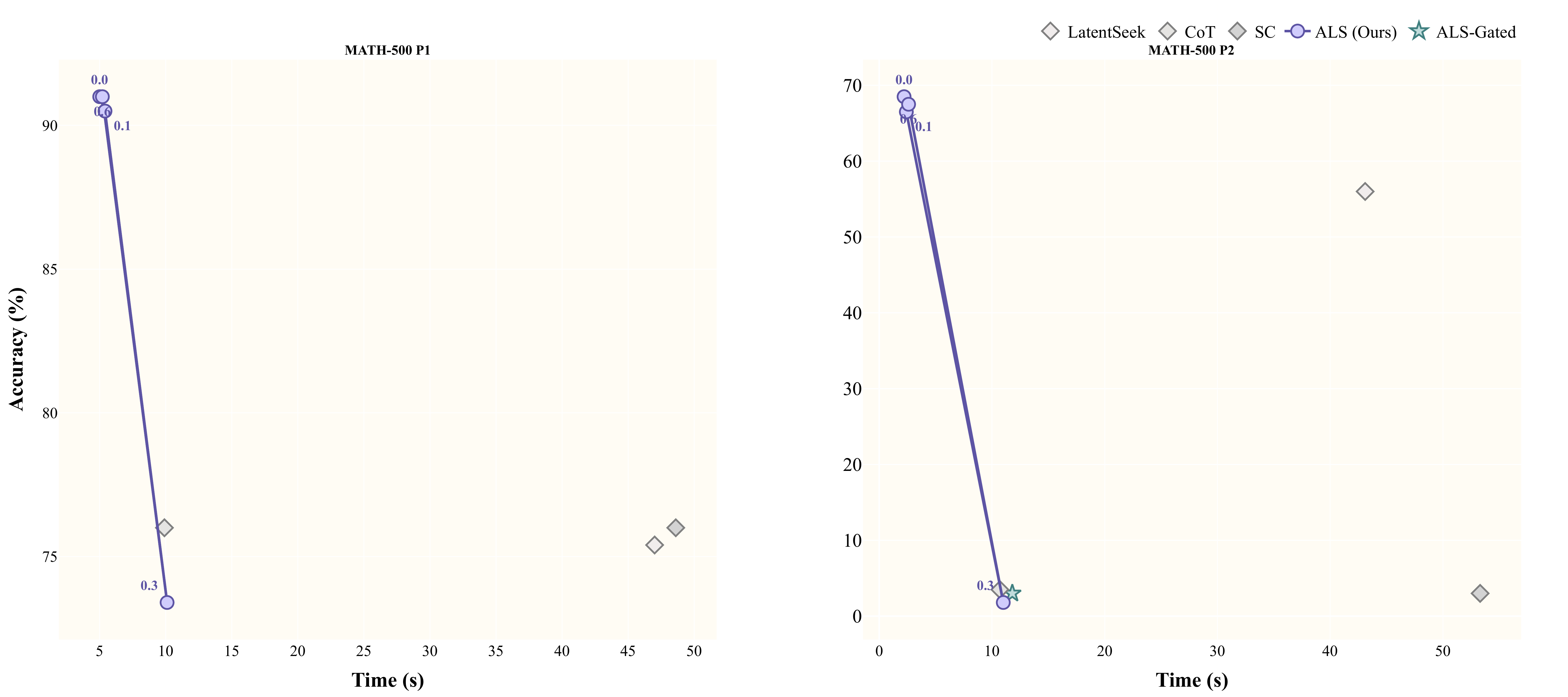}
    \caption{Pareto frontier for \textsc{ALS} on MATH-500. Results illustrate higher sensitivity to $\alpha$ compared to GSM8K, with both weak and strong steering outperforming moderate values depending on prompt style.}
    \label{fig:math500_pareto}
\end{figure}

\textsc{ALS} ablations were conducted on Qwen-2.5-7B-Instruct for efficiency; Llama runs were omitted due to computational constraints. We ablate steering strength ($\alpha$) and evaluate the gated variant. Table~\ref{tab:ablations} shows that moderate $\alpha$ values improve free-form accuracy, while weak or strong steering is better suited for structured constraints.
Figures~\ref{fig:gsm_pareto} and \ref{fig:math500_pareto} visualize the Pareto frontier across $\alpha$ values, showing the trade-off between accuracy and inference time.

\begin{table*}[h!]
\centering
\small
\caption{Ablation results for \textsc{ALS} on Qwen-2.5-7B-Instruct. We sweep steering strength $\alpha \in \{0.0, 0.1, 0.3, 0.6\}$ and report accuracy (\%), average generation time (s), and trade-off score (Equation~\ref{eq:tradeoff}). For $\alpha=0.0$, the model still uses offline training data for reasoning. Llama results are omitted due to computational constraints. The best performances are highlighted in bold.}
\label{tab:ablations}
\begin{tabular}{l l l c c c}
\toprule
Dataset & Prompt & Alpha & Accuracy (\%) & Time (s) & Trade-off \\
\midrule
GSM8K   & P1 & $\alpha=0.0$ & 76.0 & 6.8 & 84.6 \\
GSM8K   & P1 & $\alpha=0.1$ & 76.0 & 6.7 & 84.7 \\
GSM8K   & P1 & $\alpha=0.3$ & 90.6 & 5.1 & \textbf{88.8} \\
GSM8K   & P1 & $\alpha=0.6$ & 86.5 & 5.4 & 87.1 \\
\cmidrule(lr){1-6}
GSM8K   & P2 & $\alpha=0.0$ & 23.0 & 16.4 & 54.3 \\
GSM8K   & P2 & $\alpha=0.1$ & 23.3 & 16.3 & 54.5 \\
GSM8K   & P2 & $\alpha=0.3$ & 70.4 & 2.7 & \textbf{75.1} \\
GSM8K   & P2 & $\alpha=0.6$ & 70.5 & 2.0  & 74.8 \\
\cmidrule(lr){1-6}
MATH-500 & P1 & $\alpha=0.0$ & 91.0 & 5.0 & \textbf{93.1} \\
MATH-500 & P1 & $\alpha=0.1$ & 90.5 & 5.4 & 93.0 \\
MATH-500 & P1 & $\alpha=0.3$ & 73.4 & 10.1 & 76.3 \\
MATH-500 & P1 & $\alpha=0.6$ & 91.0 & 5.2 & \textbf{93.1} \\
\cmidrule(lr){1-6}
MATH-500 & P2 & $\alpha=0.0$ & 68.5 & 2.2 & \textbf{83.8} \\
MATH-500 & P2 & $\alpha=0.1$ & 66.5 & 2.4 & 83.0 \\
MATH-500 & P2 & $\alpha=0.3$ & 1.8  & 11.0 & 40.6 \\
MATH-500 & P2 & $\alpha=0.6$ & 67.5 & 2.6 & 83.2 \\
\bottomrule
\end{tabular}
\end{table*}

\subsection{Mechanism Analysis}

We hypothesize that \textsc{ALS} primarily improves performance by resolving ambiguity when the model encounters multiple competing reasoning paths. During complex reasoning, language models may explore several plausible solution trajectories in latent space, each leading to different answers. The steering vector, computed from successful versus unsuccessful trajectories, provides a directional signal that points toward the region where correct reasoning typically occurs. This effectively gives the model additional context about which path to follow when faced with conflicting options, similar to how ground truth examples guide decision-making.

By nudging hidden states toward regions historically associated with correct outputs, \textsc{ALS} helps the model commit more confidently to productive reasoning paths rather than drifting into incorrect solution spaces. This is particularly valuable on harder problems (MATH-500 vs GSM8K) and structured outputs (P2 vs P1) where the model faces more constraints and potential divergence points. \textsc{ALS} thus improves reliability while serving as a probe into latent-space geometry (see Figures~\ref{fig:gsm_pareto} and \ref{fig:math500_pareto} for visualizations).

\section{Conclusion}

We introduced \textsc{ALS}, which replaces per-input optimization with a precomputed vector derived from successful generations. On GSM8K and MATH-500, \textsc{ALS} achieves $2$--$5\times$ faster inference than LatentSeek and matches or surpasses strong decoding baselines including CoT and Self-Consistency. The method works by providing implicit guidance from successful reasoning trajectories, helping models resolve ambiguity when multiple competing solution paths are available in latent space. Like other methods, performance requires hyperparameter tuning of $\alpha$. \textsc{ALS} offers a low-cost, interpretable approach for guiding hidden representations and probing latent-space geometry.

\paragraph{Limitations.} \textsc{ALS} relies on a single global steering vector, which may not capture the full complexity of reasoning trajectories across diverse problem types. Effectiveness appears architecture-dependent, with different optimal $\alpha$ values across model families, suggesting limited generalizability. Following LatentSeek, our evaluation focuses on mathematical reasoning with two prompt formats, leaving broader domain applicability uncertain. The offline computation requires ground-truth labels for steering vector construction, which may not be available for all tasks.

\paragraph{Future Work.} Future directions include extending \textsc{ALS} beyond a single global vector to multivector or task-adaptive steering, exploring dynamic or layer-wise injection strategies for multiconstraint reasoning, and applying \textsc{ALS} to domains beyond math, such as coding or scientific QA. We also see potential in using \textsc{ALS} as a diagnostic tool for latent-space geometry to inform future model design.

\section{Broader Impacts}

\textsc{ALS} aims to improve computational efficiency of reasoning systems with several potential societal implications. Positive impacts include democratizing access to advanced reasoning capabilities through reduced computational costs and lower energy consumption at scale. However, more efficient reasoning systems could accelerate both beneficial and harmful AI applications. While our work focuses on mathematical reasoning, similar techniques could extend to other domains with dual-use implications. The method introduces no new safety concerns beyond those of the underlying language models.

\bibliographystyle{plain}
\bibliography{main}

\end{document}